\documentclass{article}

\usepackage{aaai4nmr}
\usepackage{graphicx}
\usepackage{latexsym}
\usepackage{makeidx}

\begin{document}

\newtheorem{theorem}{Theorem}
\newtheorem{definition}{Definition}
\newtheorem{example}{Example}
\newcommand{\NN}{{\rm I\!N}}

\newcommand{\dr}[3]{\ensuremath{\frac{{#1}\,:\,{#2}}{#3}}}
\newcommand{\prereq}[1]{\ensuremath{{\it Prereq}{\left( #1 \right)}}}
\newcommand{\conseq}[1]{\ensuremath{{\it Conseq}{\left( #1 \right)}}}
\newcommand{\justif}[1]{\ensuremath{{\it Justif}{\left( #1 \right)}}}
\newcommand{\theory}[1]{\ensuremath{T\!h\!\left({#1}\right)}}
\newcommand{\default}{\dr{\alpha}{\beta_1,\dots,\beta_n}{\gamma}}
\newcommand{\CupEi}{\bigcup_{i = 0}^{\infty} E_i}
\newcommand{\CupEk}{\bigcup_{k = 0}^{\infty} E_k}
\newcommand{\CupJi}{\bigcup_{i = 0}^{\infty} J_i}
\newcommand{\XRay}{\sf XRay}
\newcommand{\DefGen}{\ensuremath{DG(W,D,E)}}
\newcommand{\CExt}{\ensuremath{CE(W,D,G)}}
\newcommand{\CExtt}[1]{\ensuremath{CE_{#1}(W,D)}}
\newcommand{\CGD}{\ensuremath{CGD(W,D,G)}}

\bibliographystyle{aaai}

\title{Genetic Algorithms for Extension Search in Default Logic}
\author{Pascal Nicolas \and Fr\'ed\'eric Saubion \and Igor St\'ephan \\
LERIA, Universit\'e d'Angers \\
2 Bd Lavoisier\\
F-49045 Angers Cedex 01\\
{\small \tt \{Pascal.Nicolas,Frederic.Saubion,Igor.Stephan\}@univ-angers.fr}
}
\maketitle

\begin{abstract}
\begin{quote}
 A default theory can be characterized by its sets of plausible
  conclusions, called its extensions. But, due to the theoretical
  complexity of Default Logic ($\Sigma_2^p-complete$), the problem of
  finding such an extension is very difficult if one wants to deal
  with non trivial knowledge bases.  Based on the principle of natural
  selection, Genetic Algorithms have been quite successfully applied
  to combinatorial problems and seem useful for problems with huge
  search spaces and when no tractable algorithm is available.  The
  purpose of this paper is to show that techniques issued from Genetic
  Algorithms can be used in order to build an efficient default
  reasoning system. After providing a formal description of the
  components required for an extension search based on Genetic
  Algorithms principles, we exhibit some experimental results.
\end{quote}
\end{abstract}

\section{Introduction}
\label{sec:intro}
{\em Default Logic} has been introduced by Reiter~\cite{reiter80} in
order to formalize common sense reasoning from incomplete information,
and is now recognized as one of the most appropriate framework for
{\em non monotonic reasoning}.  In this formalism, knowledge is
represented by a default theory from which one tries to build some
extensions, that is a set of plausible conclusions.  But, due to the
level of theoretical complexity of Default Logic, the computation of
these extensions becomes a great challenge.

Previous works~\cite{chmamitr99,schaub97z,niemela95a,schris95a}
have already investigated this computational aspect of Default Logic.
Even if the system DeRes~\cite{chmamitr99} has very good performance 
on certain classes of default theories, there is no efficient system for 
general extension calculus. The aim of the present work is not to exhibit a
system able to compute extensions of every default theory in a minimal time,
but to show that techniques issued from Genetic
Algorithms can be very useful in order to build
an efficient default reasoning system.

Based on the principle of natural selection, {\em Genetic Algorithms}
have been quite successfully applied to combinatorial problems such as
scheduling or transportation problems. The key principle of this
approach states that, species evolve through adaptations to a changing
environment and that the gained knowledge is embedded in the structure
of the population and its members, encoded in their chromosomes. If
individuals are considered as potential solutions to a given problem,
applying a genetic algorithm consists in generating better and better
individuals w.r.t. the problem by selecting, crossing and mutating
them. This approach seems very useful for problems with huge search
spaces and for which no tractable algorithm is available, such as our
problem of default theory's extension search.

Here, the main difference with common uses of Genetic Algorithms is the domain
of computation. One has to point out the symbolic aspect of the search space,
since the extensions we want to compute are sets of propositional formulas. 

The paper is organized as follows : first we recall basic definitions and
concepts related to Default Logic and Genetic Algorithms.  Then, we provide the
formal description of an extension search system based on Genetic Agolrithms
principles and, at last, we describe our experiments w.r.t.  other existing
systems.

\section{Technical Background}
\label{sec:techback}
Default Logic is a non monotonic logic since the sets of conclusions (theorems)
does not necessary grow when the set of premises (axioms) does, as it is always
the case in classical logic.  In Default Logic, such a maximal set of
conclusions is called an \emph{extension} of the given default theory $(W,D)$
where $W$ is a set of first order formulas representing the sure knowledge, and
$D$ a set of \emph{default rules} (or defaults).  A \emph{default}
$\delta=\default{}$ is an inference rule providing conclusions relying upon
given, as well as absent information meaning ``if the \emph{prerequisite}
$\alpha$ is proved, and if for all $i=1,\dots,n$ each \emph{justification}
$\beta_i$ is individually consistent (in other words if nothing proves its
negation) then one concludes the \emph{consequent} $\gamma$''. 
 For a default rule $\delta$,  $\prereq{\delta}$, $\justif{\delta}$ and $\conseq{\delta}$
respectively denotes the prerequisite, the set of justifications and the
consequent of $\delta$. These definitions will be also extended to sets of
defaults.  The reader who is not familiar with Default Logic will find
in~\cite{besnard89,anton97,schaub97z} many other complements about this
formalism.  Therefore, we recall here the essential formal definitions in the
context of propositional default theories since our work is concerned by these
ones.

\begin{definition}\cite{reiter80}
  \label{def:extension}
Let $(W,D)$ be a default theory. For any set of formulas $S$ let $\Gamma(S)$
the smallest set satisfying the following properties.
\begin{itemize}
\item $W \subseteq \Gamma(S)$
\item $\theory{\Gamma(S)} = \Gamma(S)$
\item if $\default \in D$ and $\alpha \in \Gamma(S)$ and 
  $\neg \beta_1,\dots,\neg \beta_n \not\in S$, then $\gamma \in  \Gamma(S)$
\end{itemize}
A set of formulas $E$ is an extension of  $(W,D)$ iff $\Gamma(E)=E$.
\end{definition}

Based on this fixed-point definition, Reiter has given the following
pseudo iterative characterization of an extension.

\begin{definition}\cite{reiter80}
\label{def:extension_iter}
Let $(W, D)$ be a default theory and $E$ a formula set.
We define 
\begin{itemize}
\item $E_0 = W$
\item and for all  $k \geq 0$, \\
  \begin{eqnarray*}
   E_{k+1} &=& \theory{E_k}\cup \{\gamma \mid \default \in D,   \\
     && \alpha \in E_k,\neg\beta_i \not\in E,  \forall i=1,\dots,n \}
  \end{eqnarray*}

\end{itemize}
Then, $E$ is an extension of $(W, D)$ iff $E = \CupEk$.
\end{definition}

\begin{example}
\label{ex:extension}
To illustrate these definitions we give three examples, in order to describe three
particular points about default theories.
\begin{itemize}
\item  $(W_1,D_1)=(\{a, b \vee c\},\{\dr{a}{\neg b}{d}, \dr{c}{e}{e}, \dr{d}{f}{g}\})$
has a unique extension $Th(W_1 \cup \{d,g\})$.
\item  $(W_2,D_2)=(\{a, b \vee c\},\{\dr{a}{\neg b}{\neg b},\dr{a}{\neg c}{\neg c}\})$
has two extensions $E=\theory{W_2 \cup \{\neg b\}}$ and 
 $E'=\theory{W_2 \cup \{\neg c\}}$
\item  $(W_3,D_3)=(\{a\}, \{\dr{a}{b}{\neg b}\})$ has no extension.
\end{itemize}
\end{example}

As mentioned in introduction, the computation of an extension is known
to be $\Sigma_2^p-complete$~\cite{gottlob92}. Intuitively, these two
levels of complexity are due to the fact that for each default in $D$
we have to prove its prerequisite and to check that we have no proof
of the negation of one of its justification.  But, in fact, building
an extension consists in finding its \emph{Generating Default Set}
because this particular set contains all defaults whose consequents are used
to build the extension.
\begin{definition}
  \label{def:gds}
  Given $E$ an extension of a default theory $(W,D)$, the set
\[
\DefGen= \left\{ 
  \begin{array}{l}
    \default \in D \mid \alpha \in E, \\[1mm]
    \neg\beta_i  \not\in E, \forall i=1,\dots,n
  \end{array}
 \right\}
\]
is called the generating default set of $E$.
\end{definition}
Defaults that occur in the generating default set are said to be \emph{applied}
and every  generating default set is \emph{grounded}.
\begin{definition}\cite{schwind90}
  \label{def:grounded}
  Given a default theory  $(W,D)$, a set of default $\Delta \subseteq D$
  is grounded if $\Delta$ can be ordered as the following sequence 
  $<\delta_1,\dots,\delta_n>$ satisfying the property:
  \[
  \forall i=1,\dots,n, W\cup\conseq{\{\delta_1,\dots,\delta_{i-1}\}} \vdash \prereq{\delta_i}
  \]
\end{definition}

Now, we briefly recall the Genetic Algorithms concepts we use. We have to adapt
some basic techniques and modify some definitions to fit our context but we
refer the reader to \cite{micha} for a survey.

Since Genetic Algorithms are based on the principle of natural selection, vocabulary issued from natural genetics will be used in the Genetic Algorithms
framework. We first consider a {\em population} of individuals which are represented
by their {\em chromosome}. Each chromosome represents a
potential solution to the given problem. The semantics of a chromosome (called
its phenotype) has to be defined externally by the user. Then, an evaluation
process and genetic operators determine the evolution of the population in
order to get better and better individuals. 

A genetic algorithm consists of the following components :
\begin{itemize}
\item
a representation of the potential solutions : in most cases, chromosomes will
be strings of bits representing its {\em genes}, 
\item
a way to create an initial population,
\item
an {\em evaluation function} $eval$ : the evaluation function rates each potential solution
w.r.t. the given problem,
\item
genetic operators that define the composition of the children : two different
operators will be considered: {\em Crossover} allows to generate two new chromosomes
(the offsprings) by crossing two chromosomes of the current population (the
parents), {\em Mutation} arbitrarily alters one or more genes of a selected chromosome,
\item
parameters : population size $p_{size}$ and probabilities of crossover $p_c$
and mutation $p_m$.
\end{itemize}

We now present the general mechanism. Chromosomes, denoted $G_i$, are strings of bits of length
$n$. The initial population is created by generating $p_{size}$ chromosomes
randomly. Starting from this initial population, we have to define a selection
process for the next population and how to apply genetic operators. 

The selection process presented here is based on an ordering of the individuals
w.r.t. their evaluation. This process slightly differs from the initial
definition of selection in \cite{micha} which is based on the construction of a
roulette wheel by scaling.

\begin{itemize}
\item
for each chromosome $(G_i), i \in \{1..p_{size}\}$, calculate $eval(G_i)$,
\item
order\footnote{Remark that the evaluation function provides a
partial order on chromosomes which is arbitrarily extended to any total order.} the population according to evaluation rates; note that identical
individuals occur only once in this classification.
\end{itemize}

Then, an intermediate population is constructed by selecting chromosomes according to the
following method :
\begin{itemize}
\item
consider the ordered list of the different chromosomes,
\item
a decreasing number of occurrences of each chromosome is put in the selected population w.r.t. 
the place of the chromosome in this ordered list. For instance the best rated
chromosome will be represented N times in this selected population, while next
chromosome will occur N-1 times and so on... 
\item
this repartition in this population is user-defined but should satisfy that
its size is equal to $p_{size}$. 
\end{itemize}

\begin{figure}[htbp]
\includegraphics{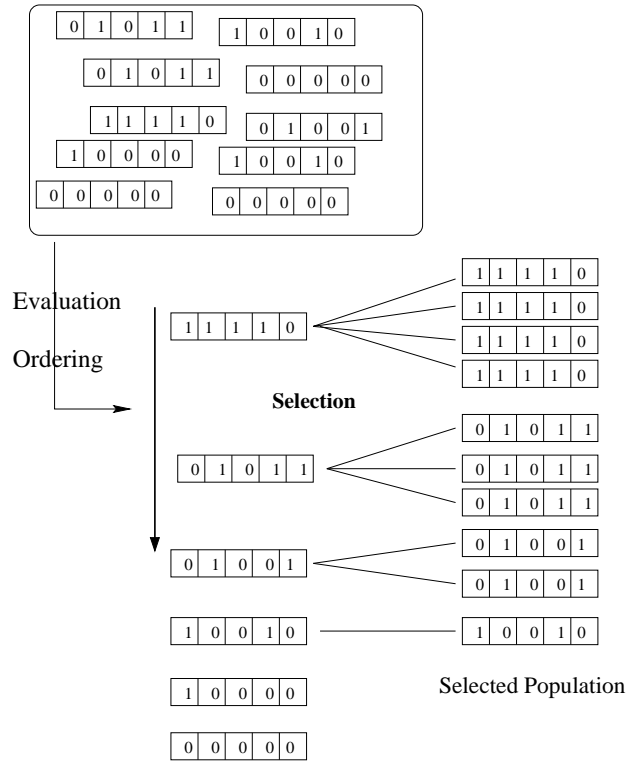}
\caption{Generation}
\label{generation}
\end{figure}

This principle is illustrated on the example of Figure~\ref{generation} where
the evaluation corresponds to the number of $1$ in the chromosome. Furthermore, the best
chromosome is duplicated 4 times in the selected population , the second 3 times,
the third 2 times and the fourth only once. Due to the extension of the order,
one can remark that, even if their rating is the same, the chromosome $(10010)$ is selected once while $(01001)$
is selected twice. This is due to the fact that $(01001)$ is greater than
$(10010)$ in the ordering. This example only shows how individuals are selected
from a population to be involved in reproduction and mutation

\begin{figure}[htbp]
\includegraphics{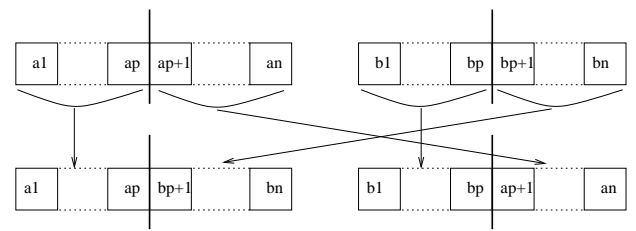}
\caption{Crossover}
\label{crossover}
\end{figure}

Therefore genetic operators will be now apply on this selected population. Crossover is performed in the
following way :
\begin{itemize}
\item
select randomly two chromosomes in the selected population
\item
generate randomly a number $r \in [0,1]$
\item
if $r>p_c$ then the crossover is possible; 

\begin{itemize}
\item 
select a random position $p \in \{ 1,\dots, n-1 \}$
\item
the two chromosomes $(a_1,...,a_{p},a_{p+1},...,a_n)$
and $(b_1,...,b_{p},b_{p+1},...,b_n)$ are replaced by the two new
chromosomes $(a_1,...,a_{p},b_{p+1},...,b_n)$ and
$(b_1,...,b_{p},a_{p+1},...,a_n)$ as shown in Figure~\ref{crossover}.
\end{itemize}
\item
if the crossover does not occur then the two chromosomes are put back in the
selected population.  
\end{itemize}

The mutation is defined as : 
\begin{itemize}
\item
For each chromosome $G_i,  i \in \{1..p_{size}\}$ and for each bit $b_j$ in
$G_i$, generate a random number $r \in [0,1]$, 
\item
if $r > p_m$ then mutate the bit $b_j$ (i.e. flip the bit).
\end{itemize}

This full process is repeated to generate successive populations and one has to
define the number of populations to be explored. The best chromosome of each
population w.r.t. the evaluation function represents the current best solution to the problem.

Clearly, the main difficulty of defining a Genetic Algorithms based search lies in
the choice of the population's representation and in the definition of the
evaluation process. A lot of work has also to be done in order to get a fine
tuning of the different parameters $p_{size},p_c,p_m$. Concerning our
particular problem, these steps will be fully detailed in the next section.

\section{Formal Description of the System}
\label{sec:desc}

Our purpose is to construct an extension of a given default theory
$(W,D)$ w.r.t. Definition~\ref{def:extension}. We call {\em candidate
  extensions} the possible solutions to our problem.  According to the
principles of Genetic Algorithms, we now consider a population of
individuals representing candidate extensions.

A naive approach could consist in considering the underlying set of atomic propositions
induced by the signature of the default theory. Thus, the chromosomes would represent a kind of truth table :

\begin{example}
With the signature ${a,b,c,d}$ an individual G,
\[
\begin{array}{l l c c c c c c c c l}
   &  & a & b & c & d & \neg a & \neg b & \neg c & \neg d & \\
G= & (& 1 & 0 & 1 & 1 & 0      &  0     &   0    &  0     & )
\end{array}
\]

represents the candidate extension $Th(\{a,c,d\})$.

It is clear that due to the basic definition of Default Logic for a default
$\dr{a}{b}{c}$ either $b$ and
$\neg b$ has to be represented in the chromosome since in Definition~\ref{def:extension} one has
to check that $\neg b \not\in S$ but this is not
equivalent to $b \in S$. Consider the following default
theory $(W,D)$ with $W = \{a\}$ and $D=\{\dr{a}{b}{c},\dr{a}{\neg b}{d}\}$. It
has only one extension $Th(\{a,c,d\})$ which does not contain $b$ neither $\neg
b$. This representation will produce a lot of inconsistent candidate extensions
because both $b$ and $\neg b$ can be marked as potentially valid as it is specified in $G$. 
\end{example}
Therefore, it seems
impossible to insure the efficiency and the convergence of the mechanism. One
solution could be to introduce a three-valued logic representation but, in this
case chromosomes cannot be strings of bits and require a more complicated encoding. 

To avoid these drawbacks, another approach consists in focusing on the defaults
more than on their consequences (according to Definition~\ref{def:extension_iter}). 
Moreover, this approach seems to be natural since an extension is completely
determined by its generating default set. 
The following definitions set out a common formal framework which consists of a
representation scheme and of an evaluation process. 

\subsection{Representation}

A representation consists of the following elements :
\begin{itemize}
\item
a chromosome language ${\mathcal G}$ defined by a chosen size $n$,
\item
an interpretation mapping to translate chromosomes in term of possibly applied defaults, which
provides the semantics of the chromosomes.
\end{itemize}

In this context, the chromosome language ${\mathcal G}$ is the regular language
$(0+1)^n$ (i.e. strings of $n$ bits). Given a chromosome $G \in {\mathcal G}$, $G|_i$ denotes the value of
$G$ at occurrence $i$. 

The mapping can be formally defined as :
\begin{definition}
Given a default theory $(W,D)$ and chromosome language ${\mathcal G}$, an interpretation
mapping is defined as : 
\[
\phi\colon {\mathcal G} \times D \to \{ true,false \}
\]
\end{definition}

A candidate extension \CExt~is
associated to each chromosome and can also be characterized by its candidate generating
default set \CGD (see Definition~\ref{def:gds}). These two sets are
easily defined w.r.t. the interpretation mapping.

\begin{definition}
Given a default theory $(W,D)$, a chromosome $G \in {\mathcal G}$, the
candidate generating default set associated to $G$ is :
$$\CGD = \{ \delta_i \mid \phi(G,\delta_i) = true \}$$
\end{definition}

\begin{definition}
Given a default theory $(W,D)$, a chromosome $G \in {\mathcal G}$, the
candidate extension associated to $G$ is :
$$\CExt = \theory{W \cup \left\{
    \begin{array}{l}
      \conseq{\delta},\\
      \delta \in \CGD 
    \end{array}
\right\}
}$$
\end{definition}

$\CExt$ and $\CGD$ will be simply denoted $CE(G)$ and $CGD(G)$ when it is clear from the
context. Remark that since we have to compute the set of logical consequences, a theorem
prover will be  needed in our system. We now comment two different possible representations according to the previous
definitions.

\begin{itemize}
\item
Given a set of defaults $D = \{ \delta_1, \cdots , \delta_n \}$ we can choose
to encode in the chromosome the fact that the default is applicable. In this
case the size of the chromosome corresponds to the cardinality of $D$
(i.e. $n$) and the interpretation function is defined as : 
\[
\forall \delta_i
\in D, \phi(\delta_i)=
\left\{
\begin{array}{l}
true \hbox{ if } G|_i=1\\
false \hbox{ if } G|_i=0\\
\end{array}
\right.
\]
The main problem with this representation is its sensitiveness to mutation and
crossover since a bit flipping in the chromosome induces a great change in the
candidate extension. To refine this, we suggest another solution.
\item
For each default $\default$ we encode in the chromosome the prerequisite
$\alpha$ and all justifications $\beta_1,...,\beta_n$ conjointly. Given a set of defaults $D =
\{ \delta_1, \cdots , \delta_n \}$ the size of the chromosome will be $2n$ and
its semantics is given by the interpretation mapping :
\[
\forall \delta_i
\in D, \phi(\delta_i)=
\left\{
\begin{array}{l}
true \hbox{ if } G|_{2i-1}=1 \hbox{ and } G|_{2i}=0 \\
false \hbox{ in other cases}\\
\end{array}
\right.
\]

Intuitively, for a default $\delta_i$, if  $G|_{2i-1}=1$ then its prerequisite
is considered to be in the candidate extension and if $G|_{2i}=0$ no
negation of its justifications is assumed to belong to the candidate
extension.This representation is chosen for the remaining of this paper. 

\begin{example}
Let consider a default theory $(W,D)$ where 
$D = \{
\dr{a}{b}{c},\dr{a}{\neg c}{\neg b},\dr{d}{e}{f}
\}$
and
$W=\{a\}$.
We get 
$CGD(100011) = \{\dr{a}{b}{c}\}$
and then
$CE(100011) = Th(\{ a,c \})$
which is really an extension but also
$CGD(101011) = \{ \dr{a}{b}{c},\dr{a}{\neg c}{\neg b}\}$
and 
$CE(101011) = Th(\{a,c,\neg b\})$
which is not an extension (negations of the justification of the two defaults
are in the set).
\end{example}
\end{itemize}
Once the representation has been settled,
one has to describe the evaluation process and then to run the genetic
algorithm principles over the population of chromosomes.

\subsection{Evaluation}

An evaluation can be defined as :
\begin{definition}
Given a chromosome language ${\mathcal G}$, an evaluation function is a mapping
$eval \colon {\mathcal G} \to {\mathcal A}$, where ${\mathcal A}$ is any set
such that there exists a total ordering $<$ on it (to achieve the selection process).
\end{definition}

Here, the evaluation function is mainly based on the definition of the
extension. Different problems can be identified providing different evaluation
criteria. 

For a default $\delta_i = \dr{\alpha_i}{\beta_i^1,...,\beta^{k_i}_i}{\gamma_i}$,
an intermediate evaluation function $f$ is defined in Table~\ref{eval}. Given the two positions $G|_{2i-1}$ and $G|_{2i}$ in the
chromosome associated to the default $\delta_i$, the first point is to
determine w.r.t. these values if this default is supposed to be
involved in the construction of the candidate extension (i.e. its
conclusion has to be added to the candidate extension or not). Then,
 we check if this application is relevant.

\begin{table}[htbp]
  \begin{scriptsize}
    \[
    \begin{array}{|c|c|c|c|c|c|}
      \hline
      \hbox{Case} & G|_{2i-1} & G|_{2i} & CE(G) \vdash \alpha_i & \exists j, CE(G) \vdash \neg
      \beta_i^j & \Pi \\
      \hline
      1 &1 & 0 & true & false & n \\
      2 &1 & 0 & true & true & y \\
      3 &1 & 0 & false & true & y \\
      4 &1 & 0 & false & false & y \\
      5 &1 & 1 & true & false & y \\
      6 &1 & 1 & true & true & n \\
      7 &1 & 1 & false & true & n \\
      8 &1 & 1 & false & false & n \\
      9 &0 & 1 & true & false & y \\
      10&0 & 1 & true & true & n \\
      11&0 & 1 & false & true & n \\
      12&0 & 1 & false & false & n \\
      13&0 & 0 & true & false & y \\
      14&0 & 0 & true & true & n \\
      15&0 & 0 & false & true & n \\
      16&0 & 0 & false & false & n \\
      \hline
    \end{array}
    \]
  \end{scriptsize}
  \caption{Evaluation}
  \label{eval}
\end{table}

A $y$ in the penality column $\Pi$ means that a positive value is assigned to
$f(G|_{2i-1},G|_{2i})$. Note that only cases 1 to 4 correspond to default
considered to be applied (i.e. such that $\phi(\delta)=true$).

\subsection{Comments on penalities}
\begin{itemize}
\item
Cases 2,3,4 :\\
The consequence $\gamma_i$ is in the candidate extension (because
$G|_{2i-1} = 1$ and $G|_{2i}=0$) while the default should not have been
applied (because either $CE(G) \not \vdash \alpha_i$ or $\exists j, CE(G) \vdash \neg
\beta_i^j$).
\item
Cases 5,9,13:\\
The consequence of the default is not in $CE(G)$ while it should since the
prerequisite of the default is in the extension and no negation 
of justifications is deducible from it.
\item
Other cases :\\
Even if the chromosome value does not agree with the generated candidate
extension, these cases can be ignored since they do not affect the extension.
\end{itemize}

At last, due to the minimality condition in the extension
Definition~\ref{def:extension} we have also to take into account the
cardinality of $CGD(G)$ (noted $card(CGD(G)$). Thus, we can define the
evaluation function as :
\begin{eqnarray*}
eval \colon {\mathcal G} &\to& \NN \times \NN \\  
eval(G) &=& (\Sigma_{i \in \{1..n\}}f(G|_{2i-1},G|_{2i}),card(CGD(G))) \\
&&\mbox{where n=card(D)}
\end{eqnarray*}

The ordering for the selection process is the lexicographic extension $(<,<)$
of the natural ordering $<$ on $\NN$.

\subsection{Correctness of the Evaluation}

We examine now what we have to do when the evaluation function
attributes a value $(0,\_)$ to a chromosome G. First, let us remark that every
candidate extension $E=\CExt$ is based on the generating default set
\CGD.
Since $eval(G)=(0,\_)$, we can easily conclude that for every default
$\default \in \CGD$ we have $\alpha \in E$ and 
$\neg \beta_i \not \in E, \forall i=1,\dots,n$.
But it is not sufficient to prove that $E$ is truly an extension of the
default theory $(W,D)$ as shown in the following counter-example.
\begin{example}
\label{ex:contrex}
  Let $(W,D) = (\emptyset,\{\dr{a}{c}{b},\dr{b}{c}{a}\})$ be a default theory and
$G=(1010)$. Then, the candidate extension is $E=\CExt=\theory{\{a,b\}}$ and $eval(G)=(0,\_)$.
But, it is obvious that $E$ is not an extension of  $(W,D)$ that has only one
extension : $\theory{\emptyset}$.
\end{example}

In fact, the counter-example~\ref{ex:contrex} illustrates that our
evaluation function does not capture the groundedness (see
Definition~\ref{def:grounded}) of the generating default set of a
candidate extension.  So, when the evaluation function gives a
chromosome with a null value, we have to check if the corresponding
generating default set is grounded.  If it is the case our following
formal result ensures that we have found an extension.  If not, the
algorithm continues to search a new candidate.

\begin{theorem}
  \label{th:verif}
  Let $(W,D)$ be a default theory, $G$ a chromosome and a candidate
  generating default set $\Delta=\CGD$.\\ 
  $eval(G)=(0,\_)$ and $\Delta$ is grounded\\ 
  \textbf{iff} \\ 
  $(W,D)$ has an extension $E=\theory{W\cup \conseq{\Delta}}$ 
of which $\Delta$ is the generating default set.
\end{theorem}

See the proof in appendix.

\subsection{Technical Improvements}

Some particular types of defaults can be treated apart to improve the system.
\begin{itemize}
\item A default $\dr{\alpha}{\beta}{\neg \beta}$ has not to be
  specifically encoded in the chromosome language and can be removed
  from the initial set of default. Since as soon as this default can
  be applied it blocks itself . One has only to check that for each
  candidate extension $CE(G)$ either $\alpha \not \in CE(G)$ or $\neg
  \beta \in CE(G)$.  Moreover, we focus on this kind of defaults
  because they are very interesting in certain cases.  For instance, a
  default $\dr{}{\beta}{\neg \beta}$ ``keeps'' only extensions that
  contain $\neg \beta$.  This property is often used in the graph
  problem encoding described in~\cite{chmamitr99}.
\item
$\delta_i = \dr{\alpha_i}{\beta^1_i...\beta^n_i}{\gamma_i}$ with $W \vdash
\alpha_i$ : then for every chromosome $G$ we
impose $G|_{2i-1}=1$.
\item
$\delta_i = \dr{\alpha_i}{\beta^1_i...\beta^n_i}{\gamma_i}$ with $W \vdash
\neg \beta^j_i$ : for some $j$ then for every chromosome $G$ we
impose $G|_{2i}=1$.\\
\end{itemize}

\section{Experimental Results :\\ the GADEL System}
\label{sec:impl}
 
\begin{center}
  \begin{figure}[htbp]
    \includegraphics{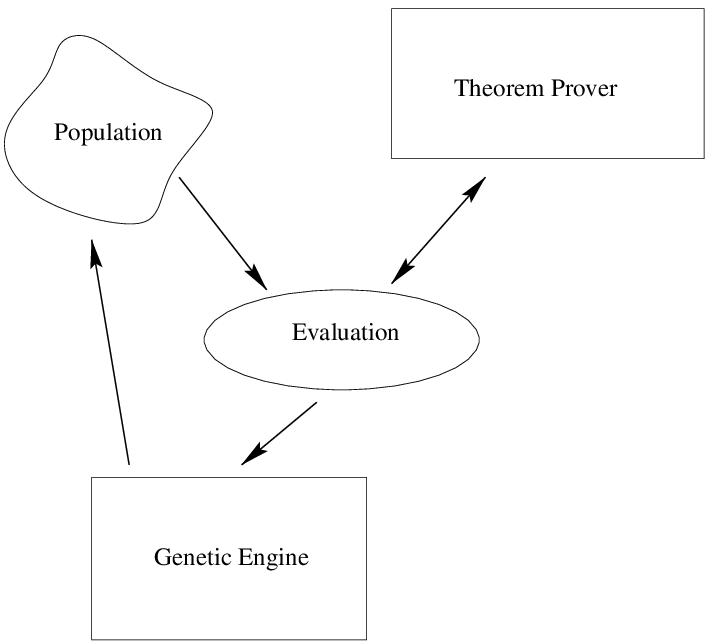}
    \caption{System}
    \label{system}
  \end{figure}
\end{center}

Our whole system GADEL (\emph{Genetic Algorithms for DEfault Logic})
can be schematized by the Figure~\ref{system}.
It is implemented in Sicstus Prolog and it is described with more
details in~\cite{stesaunic00}.

Basically, DeRes~\cite{chmamitr99} and our system GADEL use a common
approach in their search for an extension of a default theory $(W,D)$
: they both use a \emph{generate} and \emph{test} procedure. They
explore the search space $2^D$ and check if a subset $DG \subset D$
can be the generating default set of an extension of $(W,D)$. But,
DeRes explores the search space with an ad-hoc backtracking procedure
while GADEL uses the Genetic Algorithms principles in order to reach
as quickly as possible some ``good'' candidates.  ~\cite{chmamitr99}
describes the very good performances of DeRes on some kind of default
theories : the stratified ones.  But it is also noticed that for a non
stratified default theory, as for the Hamiltonian cycle problem, the
performance of DeRes are not enough to deal with a non very few number
of defaults.  

\begin{table}[htbp]
\begin{center}
\begin{tabular}{|l|r|r|r|}
    \hline
    & \multicolumn{2}{c|}{GADEL} & DeRes \\
    \hline
    problem & $NG$ & $T_G$ & $T_D$\\
    \hline
    $boy$ & 3.3 & 15.4 & >3600\\
    $girl$& 3.4 & 15.6& >3600\\
    $man$& 5.3 & 22.5& >3600\\
    $woman$ & 3.0 & 14.6& >3600\\
    $man \wedge student$& 186.7 & 467.5 & >3600 \\
    $woman \wedge student$& 271.6 & 704.4 & >3600\\
    $ham.b\_3,2,0,0,1,0,0\_$ & 1.8 &5.6 & 0.5\\
    $ham.b\_4,2,0,0,1,0,0\_$ &- &>3600 & 19.4\\
    $ham.b\_5,2,0,0,1,0,0\_$ &- &>3600 & 566.4\\
    $ham.b\_6,2,0,0,1,0,0\_$ &- &>3600 &  >3600\\
    \hline
  \end{tabular}
\caption{Experimental results}
\end{center}
\label{tab:compa}
\end{table}
 In Table~\ref{tab:compa} the first column gives the
used default theories.  For the first lines it shows the formula $f$
added to the theory $people$ (the whole description of this example is
given in appendix) and for the last ones it shows which
Hamiltonian cycle problem we have used (the encoding of the problem is
furnished by TheoryBase~\cite{chmamitr99}).  The second and third
columns respectively give average number of generations $NG$, and average
time $T_G$ in seconds to obtain one extension of $(W\cup\{f\}, D)$ by GADEL
(the parameters of the genetic algorithm are $p_c=0.8$, $p_m=0.1$,
$p_{size}=325$ for people problems, and $p_{size}=465$ for the Hamiltonian
problems, the number of tests is 100).  
The fourth column gives the time $T_D$ spent by DeRes to solve the problem with
the \emph{full prover} option. Note that all these problems are not
stratified.

We give in~\cite{stesaunic00} a finer analysis of our experiments but
results given in this table shows that DeRes has a lot of difficulties
with our taxonomic example \texttt{People} (even if we use the local
prover).  Conversely the number of generations are quite small for
GADEL (even if the time is not so good: all the implementation
is written in Prolog). But, on its turn, GADEL has poor performances
on Hamiltonian problems. We think that it is because we do not take
into account the groundedness into our evaluation function. As a
matter of fact, in the Hamiltonian problem, a solution is exactly one
``chain''\footnote{We say that $\delta$ is chained to $\delta'$ if
  $\conseq{\delta} \vdash \prereq{\delta'}$.} of defaults, but, there
is a lot of potential solutions (whose evaluation is null) based on
two, or more, chains of defaults. The only criterion to discard these
candidate generating default sets is the groundedness property that
they do not satisfy.  Conversely, in people example, a solution is a
set of non conflicting defaults, but at most four defaults are chained
together, and so the groundedness property is less important to reach
a solution.

\section{Conclusion}

The general method described in this paper provides a new framework in
order to search for extensions of a Default Logic theory, by using
Genetic Algorithms techniques. 
This new approach allows us to quickly generate
good candidate extensions and experimental results are promising w.r.t.
other systems.
Moreover, the validity of our method is ensured by a theoretical correctness
result.

Now, a first point to examine is to integrate the groundedness property in
the evaluation function, but we have to take care to not much increase the
computation time.
The efficiency could be improved by combining other search techniques
like local search heuristics. An another important feature of our
approach is its ability to be parallelized. In fact, the evaluation of
the whole population and its genetic manipulations can be
distributed on several processors without fundamental
difficulties. These points will be explored in a future work.

\appendix
\section{Appendices}

\setcounter{theorem}{0}
\subsection{Proof of the theorem} 
\begin{theorem}
  Let $(W,D)$ be a default theory, $G$ a chromosome and a candidate
  generating default set $\Delta=\CGD$.\\ 
  $eval(G)=(0,\_)$ and $\Delta$ is grounded\\ 
  \textbf{iff} \\ 
  $(W,D)$ has an extension $E=\theory{W\cup \conseq{\Delta}}$ 
of which $\Delta$ is the generating default set.
\end{theorem}

\emph{Proof} 

$\longleftarrow$: Let $E=\theory{W \cup \conseq{\Delta}}$ be an
extension of $(W,D)$.  Since $\Delta$ is the generating default set of
$E$, it is obviously grounded.  Let us suppose that $eval(G) > (0,\_)$.
Then, according to the definition of our evaluation function
(see Table~\ref{eval}), it means that there exists a default
$\delta=\default \in D$ for which a penalty has been assigned.  Let us
examine the two possible cases:
\begin{itemize}
\item  $\delta \in \Delta$: penalties can arise from cases 2, 3 or 4, but no one
of them is possible since $E\vdash \alpha$ and $E\not \vdash \beta_i, \forall i=1,\dots,n$
by definition of a generating default set
\item  $\delta \not \in \Delta$: penalties can arise from cases 5, 9, or 13, but no one
of them is possible since it would indicate that $\delta$ should be a generating default of $E$.
\end{itemize}
Thus $eval(G)=0$.

$\longrightarrow$: Let   $\Delta=\CGD$ and $E=\theory{W \cup \conseq{\Delta}}$.

Since $\Delta$ is grounded, we can order it like 
$\Delta=\langle\delta_1,\dots,\delta_p\rangle$ and we have the property
 \[
  \forall i=1,\dots,p, 
  \]
\[
W\cup\conseq{\{\delta_1,\dots,\delta_{i-1}} \vdash \prereq{\delta_i}
\]
that is equivalent to
\[
 \forall i=1,\dots,p,
\]
\[
\prereq{\delta_i} \in \theory{ W\cup\conseq{\{\delta_1,\dots,\delta_{i-1}}}
\]
from which we can build the sequence
\begin{eqnarray*}
  E_0&=&W\\
  E_{i+1}&=&\theory{E_i} \cup \{\conseq{\delta_i}\},\forall i=0,\dots,p-1   
\end{eqnarray*}
Because of the groundedness of $\Delta$, there is no difficulty to transform the previous
sequence in the following way.
\begin{eqnarray*}
  E_0&=&W\\
  (*)E_{i+1}&=&\theory{E_i} \cup \{\conseq{\delta_i} | \prereq{\delta_i} \in E_i\},\\
&&\forall i=0,\dots,p-1 
\end{eqnarray*}

Since $eval(G)=(0,\_)$, we can deduce :
\[
\forall \beta^j_i \in \justif{\delta_i}, \neg \beta^j_i\not\in E
\]
and then we can reformulate $(*)$ like that
\begin{eqnarray*}
  E_0&=&W\\  
 (**)E_{i+1}&=&\theory{E_i} \cup \{\conseq{\delta_i} | \prereq{\delta_i} \in E_i,\\
 && \beta^j_i  \in \justif{\delta_i}, \beta^j_i \not\in E \},\\
 &&\forall i=0,\dots,n-1
\end{eqnarray*}
From $eval(G)=(0,\_)$ we can also deduce that for all other defaults $\default \in D \setminus \Delta$,
we have either $\alpha \not \in E$, either $\exists j, \neg \beta_j \in E$.
So, in $(**)$ we can delete the explicit reference to $i$ in the defaults and 
we can extend the sequence for all positive integer. So we have
\begin{eqnarray*}
  E_0&=&W\\
 E_{k+1}&=&\theory{E_k} \cup \{\conseq{\delta} | \prereq{\delta} \in E_k,\\
 && j\in \justif{\delta}, j\not\in E \},\forall k>0
\end{eqnarray*}

Finally, let us remark that by construction $E$ is exactly the set $\CupEk$.
Thus we have obtain here the pseudo iterative characterization of an extension given
in Definition~\ref{def:extension_iter}, and we can conclude that $E$ is an extension
of $(W,D)$. \hfill  $\Box$

\subsection{People example}
\label{app:people}
This is the description of our examples \\
$\{boy | girl | man | woman | man \wedge student | woman \wedge student \}\_people$

\textbf{formula set}
$W = \{\neg boy \vee \neg girl$, $ \neg boy \vee kid$, $ \neg girl \vee kid$,
 $\neg human \vee male \vee female$, 
 $\neg kid \vee human$, $ \neg student \vee human$,
 $\neg adult \vee human$, $ \neg adult \vee \neg kid$,
 $\neg adult \vee \neg male \vee man$, $ \neg adult \vee \neg female \vee woman$,
 $\neg academic \vee adult$, $ \neg academic \vee diploma$,
 $\neg doctor \vee academic$, $ \neg priest \vee academic$,
 $\neg prof \vee academic$, $ \neg bishop \vee priest$,
 $\neg cardinal \vee bishop$, $ \neg redsuit \vee suit$,
 $\neg whitesuit \vee suit$, $ \neg blacksuit \vee suit$,
 $\neg redsuit \vee \neg whitesuit$, $ \neg whitesuit \vee \neg blacksuit$,
 $\neg redsuit \vee \neg blacksuit \}$\\
 $\cup \{boy\}$ or $\cup \{girl\}$ or $\cup \{man\}$ or $\cup \{woman\}$ or
 $\cup \{man, student\}$ or $\cup \{woman, student\}$

\textbf{default set} 
$D =\{ \dr{human}{name}{name}$, $ \dr{kid}{toys}{toys }$, 
$ \dr{student}{adult}{adult}$, $ \dr{student}{\neg employed}{\neg employed}$, 
$ \dr{student}{\neg married}{\neg married}$, $\dr{student}{sports}{sports}$, 
$ \dr{adult}{\neg student}{employed}$, $ \dr{adult}{\neg student, \neg priest}{married}$, 
$ \dr{adult}{car}{car}$, 
$ \dr{adult}{\neg academic}{\neg toys}$, $ \dr{man}{\neg prof}{beer}$, 
$ \dr{man}{\neg vegetarian}{steak}$, $ \dr{man}{coffee}{coffee}$,
$ \dr{man \vee woman}{wine}{wine}$, $ \dr{woman}{tea}{tea}$, 
$ \dr{academic}{\neg prof}{\neg employed}$, $ \dr{academic}{\neg priest}{toys}$, 
$ \dr{academic}{books}{books}$, $ \dr{academic}{glasses}{glasses}$, 
$ \dr{academic}{\neg priest}{late}$, $ \dr{doctor}{medicine}{medicine}$, 
$ \dr{doctor}{whitesuit}{whitesuit}$, $ \dr{prof}{employed}{employed}$, 
$ \dr{prof}{grey}{grey}$, $ \dr{prof}{tie}{tie}$, 
$ \dr{prof}{water}{water}$, $ \dr{prof}{conservative}{conservative}$, 
$ \dr{priest}{male}{male}$, $ \dr{priest}{conservative}{conservative}$, 
$ \dr{priest}{\neg cardinal}{blacksuit}$, $ \dr{cardinal}{redsuit}{redsuit}$,
$ \dr{car}{mobile}{mobile}$, $ \dr{tie}{suit}{suit}$, 
$ \dr{wine \wedge steak \wedge coffee}{\neg sports}{heartdisease}$, 
$ \dr{sports}{man}{football \vee rugby \vee tennis}$, 
$ \dr{sports}{woman}{swim \vee jogging \vee tennis}$, 
$ \dr{toys \wedge (football \vee rugby)}{ball}{ball}$, 
$ \dr{toys}{boy}{weapon}$, $ \dr{toys}{girl}{doll} \} $

\bibliography{/home/info/helios/pn/Tex/biblio,/home/info/helios/pn/Tex/bibmoi,/home/info/helios/saubion/bibliographies/biblio}

\end{document}